# A TEST BENCH FOR EVALUATING EXOSKELETONS FOR UPPER LIMB REHABILITATION

Clautilde Nguiadem[1,2*], Maxime Raison[1,2], Sofiane Achiche[1]

1 Department of mechanical engineering, École Polytechnique de Montréal, Montréal, QC, Canada

2 Technopole in pediatric rehabilitation of Ste-Justine UHC, Montréal, QC, Canada

**\* Correspondence:**

Corresponding Author

clautildenguiadem@polymtl.ca

## ABSTRACT

Rehabilitation exoskeletons have proven to be useful in assisting clinicians in therapy and assisting users in daily tasks. While the potential of wearable robotics technology is undeniable, quantifying its value is difficult. As a result, performance in wearable robotics is becoming a pressing concern, and the scientific community requires reliable and repeatable testing methodologies to evaluate and compare the available exoskeletal systems. Various types of exoskeleton robots have already been developed and tested for upper limb rehabilitation. The problem is that evaluations are not standardized, particularly in pediatric rehabilitation.

This paper aimed to propose a methodology for the quantitative evaluation of upper limb exoskeletons that, like a test bench, would serve for replicable testing. This was accomplished by determining the range of motion (ROM) and joint torques using both kinematic modeling and experimental measurements (using sensors integrated into Dynamixel actuators, where ROM and joint torques were estimated from actuator feedback, respectively, in position and in load through the IDE Arduino).

The proposed test bench can provide an accurate range of motion (ROM) and joint torques during the pronation–supination task. The range of motion obtained with the 3D or physical prototype was approximately 156.26 ± 4.71° during the pronation–supination task, while it was approximately 146.84 ± 14.32° for the multibody model. The results show that the average range of experimental torques (0.28 ± 0.06 N.m) was overestimated by 40% and just 3.4%, respectively, when compared to the average range of simulated torques (0.2 ± 0.05 N.m) and to the highest range of simulated torques (0.29 N.m). For the experimental measurements, test–retest reliability was excellent ($\alpha$ = 0.96-0.98) within sessions and excellent or good ($\alpha$ = 0.93 and $\alpha$ = 0.81-0.86) between sessions.

Finally, the suggested approach provides a range of motion close to the normal range of motion necessary during PS tasks. These results are important because they validate the measurements' accuracy and underline the proposed methodology's relevance. This study also confirms fluctuations in torque in human joints during motion and emphasizes the importance of considering these variations for precise quantification of joint torques by using the maximum value of estimated torques (rather than the average value).

To conclude, the proposed assessment procedure could become a reference standard for evaluating exoskeletons for the upper limb. This study also addresses a methodological aspect on the accurate assessment of joint torques that can serve in applications such as the sizing of actuators in exoskeletons or the non-invasive evaluation of muscle forces in the human body. In perspective, the concept will be expanded to additional joints, such as the elbow and wrist, to have a more complex assessment tool. Furthermore, future research will address the user's safety by quantifying the kinematic coupling between the user and the device.

**Keywords: upper limb rehabilitation, multibody modeling, exoskeleton evaluation, movement simulation, rehabilitation devices, motor sizing, wearable robotics.**

**Abbreviations:**

AC: acromioclavicular

ADL: activities of daily living

CAD: computer-aided design

CCW: counter-clockwise

CW: clockwise

DoF(s): degrees of freedom

FE: flexion–extension

GH: glenohumeral

HU: humeroulnar

HR: humeroradial

MBD: multibody

PS: pronation–supination

RC: radiocarpal

ROM: range of motion

RU: radioulnar

SC: sternoclavicular

SCI: spinal cord injury

SD: standard deviation

SIP: segmental inertial parameters

## 1. INTRODUCTION

In the early 1990s, the use of robotic devices in neurorehabilitation was proposed to provide motor training and to support clinicians for physical therapy [1-3]. Rehabilitation robots were developed and clinically used for patients with spinal cord injuries (SCI) [4-6], for post-stroke rehabilitation [7-9], as well as for patients with other neurological or physiological conditions, such as multiple sclerosis [10], that result in movement impairment [11-17]. Motor impairment, which usually affects the limbs, is the most common consequence/deficit among individuals after stroke and traumatic brain or spinal cord injury. The rehabilitation process restores the functionality of the upper extremities, allowing the patient to recover the greatest possible use of their limbs and regain their independence. Some of the rehabilitation modalities include electromyography biofeedback, robot-assisted therapy, virtual reality-based interventions, and functional electrical stimulation [18]. Physical rehabilitation is still crucial for individuals to regain functional independence after motor impairment. Therefore, physical therapy and exercise help to increase motor recovery and function, thanks to changes in cortical reorganization caused by residual neuroplasticity [19, 20]. However, exercise-based treatments represent a considerable burden for therapists and are heavy consumers of healthcare resources since a therapist performs the repetitive movements of the impaired limb throughout the therapeutic training session. As a result, not only does the entire rehabilitation process necessitate the presence of a professional therapist, but it is also a labor-intensive, time-consuming, and costly procedure for both therapists and patients [21]. Furthermore, the rehabilitation field faces a shortage of therapists as the population with rehabilitation needs, such as stroke patients and the elderly, grows [22]. Zimbelman et al. have reported that shortages are expected to continue to increase until 2030 in the United States [23]. To address these challenges, rehabilitation robots have been designed for use as assistive robotic devices (assistive and therapeutic robots) for therapists in the clinical setting and for patients at home. Clinical evidence and a growing number of studies have confirmed the potential benefits of robot-assisted therapy in upper limb rehabilitation [6, 24-27] and found it to be superior to standard manual therapy in some cases [15, 25, 28, 29]. The key benefits of rehabilitation robots are that they can give high-dose and high-intensity training through high-quality and repeated motions. Thus, these devices provide intensive, accurate, quantitative [30, 31], and safe rehabilitation [6]. Overall, these rehabilitation devices may help improve the upper limb rehabilitation treatments by supporting and guiding the patients to perform motor tasks while completely utilizing their residual sensorimotor coordination abilities.

The most fundamental block of a rehabilitation robot's design is its kinematic structure, which determines its functional capabilities. According to the kinematic or mechanical structure, end-effector-based systems and exoskeleton-based systems are the two major types of rehabilitation robots [32] [11, 12] [9]. The first category covers simpler devices with one or two degrees of freedom (DoF) that can be used to train basic functions or single-articulation movements (such as elbow or hand planar movements), while the second category represents multi-DOF robots that can train spatial and more complex movements. In the latter option, exoskeletons are the most advanced robots since they control the end effector of the human arm (i.e., the hand or wrist) and the entire kinematic chain, providing single-joint robotic assistance during movement execution. As a result, they can be tailored to each patient's specific needs.

While very promising, on the one hand, the development of rehabilitation exoskeletons faces design challenges, including issues related to the exoskeleton kinematic compatibility with human upper limbs [33, 34]. Due to the strong link between human limbs and exoskeleton robots, the user can be injured if the exoskeleton tries to impose incompatible kinematic or dynamic configurations on the human body [35, 36]. This is critical in motor rehabilitation because individuals may have muscle weakness. Recent rehabilitation robots are built to operate at very low power in order to minimize injury to the end user [16], but this method causes the robots to work very slowly, limiting the activities that they can perform. To better minimize such risks, the exoskeleton mechanism's safety should be built-in, and the interface should be more effective. Upper limb exoskeletons (ULE) are designed to be kinematically compatible with the wearer's joints, allowing them to undertake therapy exercises safely and efficiently. Due to the proximity between the two systems (users and robots), safety is becoming a fundamental criterion in the design of exoskeleton robots [26, 37, 38]. Safety involves risks related to the devices' use (such as the effects of the devices on motor function) as well as their regulation [39]. On the other hand, although exoskeletons represent a promising technology, many questions concerning effective robotic upper limb rehabilitation remain unanswered [12]. In fact, there is not yet sufficient evidence of the clinical effectiveness robot-assisted rehabilitation [26, 40]. This gap has been identified in the literature as resulting from the following factors: (i) the difficulty of comparing the few clinical results available; (ii) the lack of objective assessment tools; and (iii) the most effective methods are still not clear and cannot be implemented in rehabilitation robots. In addition to safety, defining the benefits of these robotic devices is critical because it clearly identifies which functions in daily life can be aided and accomplished. Robotic rehabilitation systems are used when the benefit outweighs the risk. In this context, rehabilitation robots should be evaluated on the basis of their safety and effectiveness [16], which necessitates a thorough assessment of these robots.

During the last two decades, there has been a tremendous effort to enhance the design and control strategy of robotic rehabilitation devices, but little has been done to verify their efficacy in rehabilitation settings. Many rehabilitation robots have been developed over the last decade. However, only a handful have become commercially available, and some are still in the development stage, undergoing clinical studies to gather data on their usefulness and effectiveness [12]. It has been reported that results of the clinical evaluation of therapy-applying robots are still sparse [12]. In addition, the results of controlled clinical trials on efficacy and effectiveness remain limited, and those already available are difficult to compare with one another [7, 12, 41]. As a result, quantitative evaluations of these devices are not standardized [40, 42]. Although wearable robotic technology has undeniable potential, quantifying its worth is difficult. In the field of wearable robotics, performance evaluation is becoming increasingly important [43], but it is challenging. There is currently no systematic framework for evaluating these robots in all of their aspects, particularly in pediatric rehabilitation, where access to exoskeletons lags far behind that of adults [44]. Previous reviews have noted weakness and difficulty in providing reliable evidence of these devices' clinical usefulness [12], potentially due to a lack of clear and rigorous assessment methods [12, 45]. To address this gap, quantitative benchmarking approaches are recommended; thus, evaluation of the performance and conducting appropriate testing are achieved by using a test bench [45-47]. Quantitative assessment as evaluation for rehabilitation robot benchmarks not only allows for the verification and comparison of different devices but also defines R&D targets and directions and provides important support for the standardization and efficient transfer of wearable exoskeletons from the lab to the market [12, 44, 47].

It is necessary to develop a standard benchmarking framework for wearable robots in order for them to be effectively and extensively adopted by end users [38, 42]. In this context, we provide a test bench for the evaluation of exoskeletons for upper extremity rehabilitation in this work. This test bench uses a prototype of a biofidelic prosthesis with a unique design feature: the forearm's kinematic structure is similar to that of a human [48]. To assess the device's effectiveness, we design an evaluation protocol that includes flexion–extension and pronation–supination tasks that are representative of upper limb tasks involving elbow, forearm, and wrist joint motions to assess the device's effectiveness. The evaluation of robotic-based rehabilitation devices, as with any other wearable robotic technology, should take into account two key domains: i) the functional perspective and ii) the interaction perspective [46]. In this paper, we look at the first viewpoint, namely the functional perspective, in which benchmarks are divided into two categories: performance and biomechanics. The scientific community urgently needs reliable and repeatable testing methodologies to validate and compare the performance of the many and varied exoskeletal options available [12, 29, 37, 42, 43, 49]. According to a discussion between researchers and stakeholders during the 2019 edition of the ExoBerlin conference [50], the key aspects in the field of performance evaluation to be addressed in the near future include the following: functional performance (i.e., measuring physical performance relevant to wearable exoskeletons' functioning) and methodological aspects (i.e., creating a methodology for measuring physical performance relevant to wearable exoskeletons) [46]. As a result, the primary goal of this research is to provide a test bed/test bench for the precise quantification of the performance of the proposed prototype and, as a result, to establish a standard for evaluating upper limb exoskeletons.

## 2. MATERIALS AND METHODS

A test bench based on a biofidelic upper limb prosthesis was built in Solidworks and then 3D-printed into a physical prototype in order to achieve reproducibility for the benchmarking presented in this study. We identified two essential views for the benchmarking approaches in rehabilitation robotics—functional and interaction perspectives—in order to provide a systematic framework for the evaluation of upper limb wearable exoskeletons. The functional perspective, which incorporates performance and biomechanical benchmarks, is the subject of this paper. As a result, the physical prototype of the prosthesis is presented first in this section, followed by the simulation approach of the multibody model suggested in our prior paper [48] in order to evaluate the performance of the final design (here, the kinematic data acquisition is outlined). The experimental setup and protocol for the experimental testing, as well as the methodological approach for the quantitative assessment (estimation) of performance metrics obtained from experiments, are then described. Finally, the statistical tests used in this research are described.

## 2.1 System Design

Since the upper limb exoskeleton works in constrained motion with the human body, appropriate kinematic constraints should be satisfied. The prosthesis in this work is based on a biofidelic kinematic structure of the forearm [48]. The prosthesis' kinematic structure has four degrees of freedom, supporting the motion of the elbow and wrist joints. The computer-aided design (CAD) model of the prosthesis, created with the Solidworks software, is shown in Figure 1. It weighs 3.5 kg and comprises the following components:

- Humerus, which stands for the arm part.
- Forearm section, which is represented by two rigid bodies (ulna and radius).
- Link shaft, which permits the two forearm parts to be linked together.
- Four Dynamixel AX-18A servomotors, placed at the joint level for the execution of the required movements.

The system can perform upper limb motions such as elbow flexion–extension (FE), forearm pronation–supination (PS), flexion–extension, and radio-ulnar deviation of the wrist thanks to its four degrees of freedom. Dynamixel smart servomotors were controlled using an Arduino platform algorithm to offer the desired functionality.

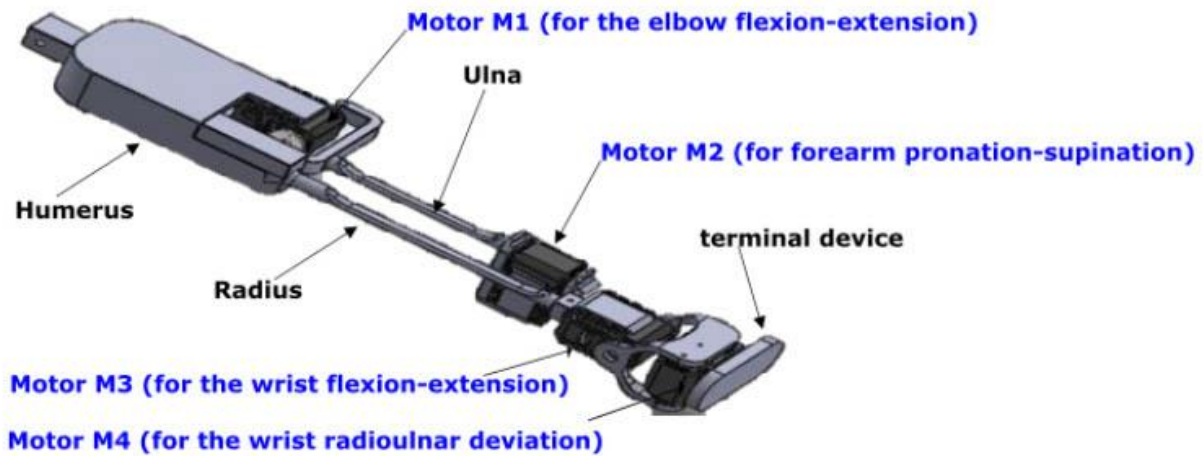

Figure 1 A Computer-aided Design (CAD) model of the upper limb prosthesis built in Solidworks

## 2.2 Multibody dynamics simulation of the prosthesis

This subsection presents the modeling and dynamic simulation of the prosthetic mechanism used in this study (i.e., the mechanism designed using the Solidworks software and printed in 3D). The kinematics and the approximate dynamic behavior of the prosthesis were estimated using multibody dynamics analysis and were then compared to experimental measurements in the rest of the study (see Section 2.3).

The equations of motion were symbolically generated using the ROBOTRAN software, using a recursive Newton–Euler formalism [51] employing relative generalized coordinates **q** as system configuration parameters. The symbolic equations were then used in Matlab to compute inverse dynamics and kinematics. These dynamic and kinematic computations were performed using Eq.(1), a motion equation formulated by using constrained Lagrange equations. The computer simulation based on multibody dynamics requires actual joint kinematics representing the input data. In a previous study of our research group [52], these kinematic data (i.e., input data) were recorded during pronation–supination (PS) and pure flexion–extension (FE) movements of 15 healthy adults by a 12-camera motion capture system (T40S, Vicon-Oxford, UK) sampled at 100 Hz.

First, from actual joint positions (**Xexp**, m) obtained from kinematic markers, the joint positions predicted by the multibody model (**Xmo**d, m *(q)*) were reconstructed by the inverse kinematics process by using the global optimization approach specified by Equation (6). The *q* variables represent joint angles, which are used to approximate the range of motion (ROM) assessed in this study. Joint torques (***Q***) required for each of the studied tasks (flexion–extension and pronation–supination) were then computed using segmental inertial parameters, and generalized positions, speeds, and accelerations $(q, \dot{q}, \ddot{q})$ using inverse dynamics (see Equation 2). Finally, the multibody dynamics simulation also provided the temporal evolution of power consumption. The power formula (**P**), as well as its relationship with total energy (**E**) as a function of angular velocity (**Ω**), are described in equations 7 and 8, respectively.

The multibody model was created using data from the real prototype of the prosthesis (i.e., mass, centers of gravity, and length of arm, forearm, and hand).

## 2.2.1 Segmental inertial parameters (SIPs)

In biomechanics, SIPs are key parameters for studying the dynamics of human motion [53]. As in our previous study [48], SIPs of the arm, the forearm, and the hand were defined by using the Yeadon's model [54] but based on the measurements (lengths and masses) of the physical prototype. Otherwise, the SIPs of the thorax, clavicle, and scapula were estimated using measurements obtained from medical imaging obtained from the literature, and undefined or unavailable values were set to negligible values.

$$\mathbf{M}(q, \delta)\,\ddot{q} + C(q, \dot{q}, \delta, frc, trq, g) = Q(q, \dot{q}) + J^T \lambda \quad (1)$$

$$Q = Q(q, \dot{q}, \ddot{q}) \quad (2)$$

$$\mathbf{h}\,(q) = h_{loop}\,(q) = 0 \quad (3)$$

$$\dot{h}(q, \dot{q}) = J(q)\,\dot{q} = 0 \quad (4)$$

$$\ddot{h}(q, \dot{q}, \ddot{q}) = J(q)\,\ddot{q} + \dot{J}(q, \dot{q})\,\dot{q} = 0 \quad (5)$$

where **M** is the matrix of the generalized system inertia, **C** is the vector of dynamic nonlinear effects containing gyroscopic, centrifugal, gravitational, and external force effects, **Q** is the vector of generalized forces, $J$ ($J = \frac{\delta_h}{\delta_q^T}$) is the Jacobian matrix of kinematic constraints $h(q)$, $\lambda$ is the vector of the Lagrange multipliers connected to the kinematic constraints, g is the gravity constant, frc and trq are external forces, and $\delta$ is the dynamic parameters of the multibody system.

Equation (1) is subjected to kinematic constraints $h(q)$, defined in this study by loop-closure constraints $h_{loop}$ (Equation 3). The geometric constraints imposed by the cut of the ball joint (HR) are checked at any time using equations 3 to 5 to ensure that the loop closure is respected.

$$\min_{q} f(q) = \sum_{m=1}^{n_m} \| X_{mod,m}(q) - X_{exp,m} \|^2 \qquad (6)$$

where q is the desired variable during the optimization, and m is the index of the marker ($n_m = 29$). f(q) is the objective function solved independently at each time of frame f, and $n_f$ indicates the total number of frames.

$$P = Q\Omega \qquad (7)$$

$$E = \int P(t)\, dt = \int |Q\Omega|\, dt \qquad (8)$$

### 2.2.2 Multibody model of the prosthesis

In a recent work, we developed a multibody (MBD) model of the upper limb prosthesis [48]. As shown in Figure 2, the kinematic chain is composed of four main parts: the shoulder complex, the arm, the forearm, and the terminal device.

Unlike the physical prototype, the multibody model included the shoulder, represented by four bodies in series (thorax, clavicle, scapula, and humerus) connected by three successive spherical joints: the sternoclavicular (SC), acromioclavicular (AC), and glenohumeral (GH) joints defined as $q_{7-9}$, $q_{10-12}$, and $q_{13-15}$, respectively. The thorax was the moving base of the MBD kinematic chain, with six degrees of freedom (DoFs) ($q_{1-6}$). The arm was represented by a single body that ran from the shoulder to the elbow and corresponded to the humerus. A parallel mechanism combining two solids, the radius and the ulna, represented the forearm. By a closed-loop PS mechanism ($q_{17-21}$) with two bodies, the forearm was attached to the humerus by the elbow joint ($q_{16}$) and to the hand by the wrist joint ($q_{22}$ - $q_{23}$) on the proximal and distal sides, respectively. Finally, the hand was modeled as a rigid body.

Roughly, the model's kinematic chain was characterized by 7 rigid body segments with 23 DoFs coupled by 9 joints, as described in Table 1. Our prior study [48], however, provides a more detailed description of the kinematic chain.

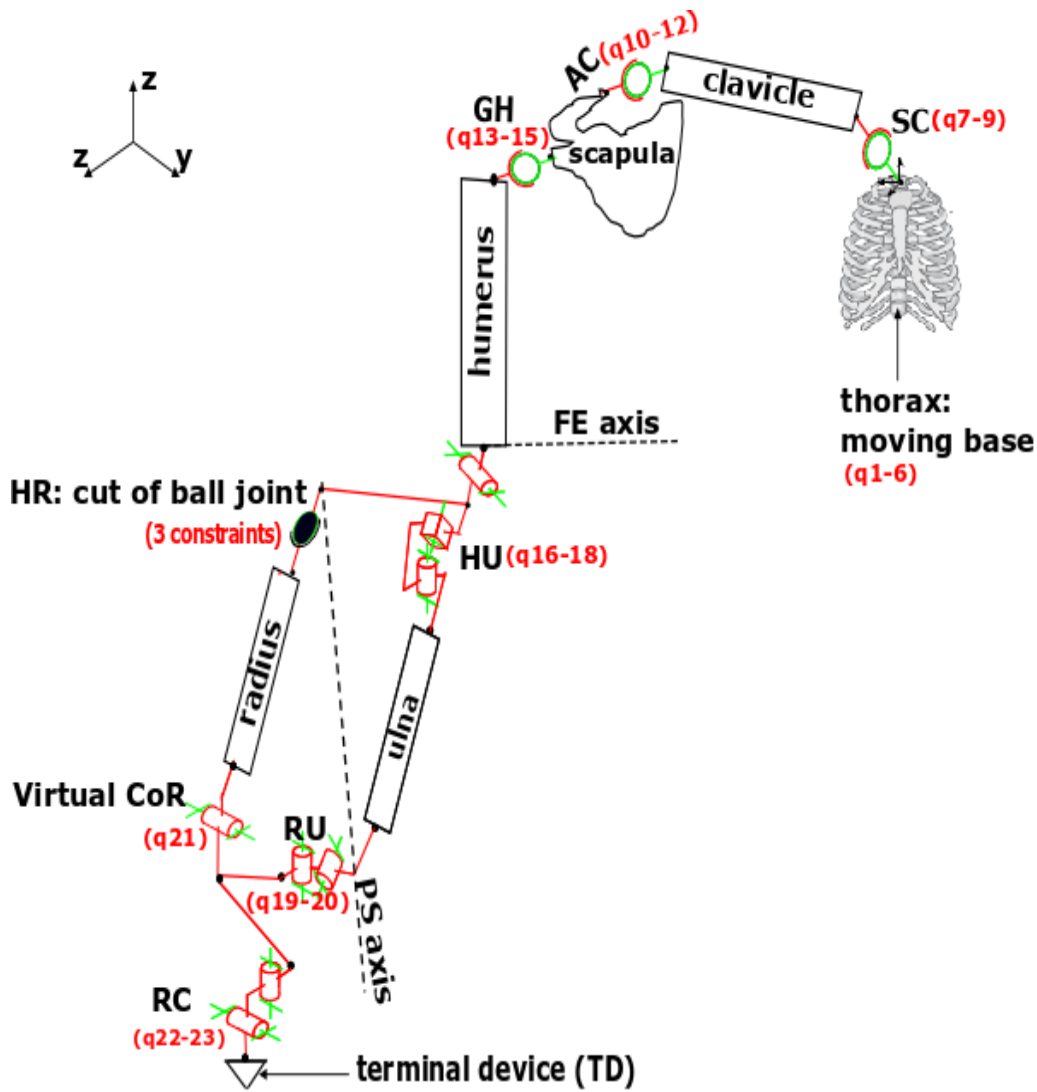

Figure 2 Multibody model of the prosthesis. The model is articulated by a moving base (q1–6), the sternoclavicular joint (SC, q7–9), the acromioclavicular joint (AC, q10–12), the glenohumeral joint (GH, q13–15), the humeroulnar joint (HU, q16–18), the radioulnar joint (RU, q19–20), the virtual

CoR (q21), the humeroradial joint (HR, cut of ball joint with three kinematic loop-closure constraints), and the radiocarpal joint (RC, q22–23). See Table 1 for the detailed description of the kinematic chain.

Table 1 kinematic chain description of the proposed model

| Joint | Proximal segment | Distal segment | DoFs | qi | Local axis | Functional description |
|---|---|---|---|---|---|---|
| Moving base | Base | Thorax | 6 | $q_1$ | Z | Medial/lateral translation |
| | | | | $q_2$ | Y | Vertical translation |
| | | | | $q_3$ | X | Anterior/posterior translation |
| | | | | $q_4$ | Z | Flexion/extension |
| | | | | $q_5$ | X | Lateral rotation |
| | | | | $q_6$ | Y | Axial rotation |
| Sternoclavicular (SC) | Thorax | Clavicle | 3 | $q_7$ | Y | Protraction-retraction |
| | | | | $q_8$ | X | Depression/elevation |
| | | | | $q_9$ | Z | Axial rotation |
| Acromioclavicular (AC) | Clavicle | Scapula | 3 | $q_{10}$ | Y | Protraction/retraction |
| | | | | $q_{11}$ | X | Lateral/medial rotation |

| | | | | $q_{12}$ | Z | Anterior/posterior tilt |
|---|---|---|---|---|---|---|
| Glenohumeral (GH) | Scapula | Humerus | 3 | $q_{13}$ | Y | Plane of elevation |
| | | | | $q_{14}$ | X | Negative elevation |
| | | | | $q_{15}$ | Y | Axial rotation |
| Humeroradial (HU) | Humerus | Ulna | 1 | $q_{16}$ | Z | Flexion/extension |
| | | | Closed-loop: 5-3=2 DoFs | $q_{17}$ | X | Axial displacement |
| | | | | $q_{18}$ | Y | Lateral swaying |
| Radioulnar (RU) | Ulna | Radius | | $q_{19}$ | X | Pronation/supination |
| | | | | $q_{20}$ | Y | Torsional angle |
| Virtual CoR | Ulna | Radius | | $q_{21}$ | Z | Aperture angle |
| Humeroradial (HR) | Radius | Humerus | | - | | Cut of ball joint with 3 constraints |
| Radiocarpal (RC) | Radius | Hand/terminal device | 2 | $q_{22}$ | Y | Flexion/extension |
| | | | | $q_{23}$ | Z | Ulnar/radial deviation |

## 2.3 Experimental tests

Following the multibody analysis, the experimental tests were carried out on the physical test bench, which is controlled by an algorithm on the Arduino platform and operated by Dynamixel smart servomotors. These experiments aimed to assess the system's performance thanks to the sensors embedded inside the servomotors. Two parameters, the range of motion (ROM) and the joint torques required to perform the studied pronation–supination (PS) movement, were chosen as performance evaluation metrics in this study, based on the definition of performance as "the level of accomplishment of a defined motor skill/task".

These parameters were measured using the experimental setup and procedure described above. To begin, we measured the range of motion (ROM), which was then compared to the simulated ROM from the multibody (MBD) model (as well as the healthy reference). Second, the torques provided by actuators at the joints were measured and compared to the torques simulated by the MBD model.

### 2.3.1 Experimental setup and procedure

The experimental setup comprised the following components, as indicated in Figure 3:

- Physical test bench: 3D-printed upper limb prosthesis (based on a compliant kinematic structure of the forearm) with Dynamixel AX-18A servomotors installed at the prosthetic joints.
- A laptop/PC, on which the Arduino software was installed and executed.
- A 9V-12V power supply was required to power the Dynamixel AX-18A actuators.
- Arduino board with the ATMEGA328p microcontroller
- USB to SERIAL cable: for data transmission and Arduino board power.
- Communication circuit (by half-duplex UART protocol) based on the 74LS241 chip, shown in Appendix 1: for communication between the microcontroller and the servomotors.
- Workstation/work surface: to set up all of the necessary equipment.

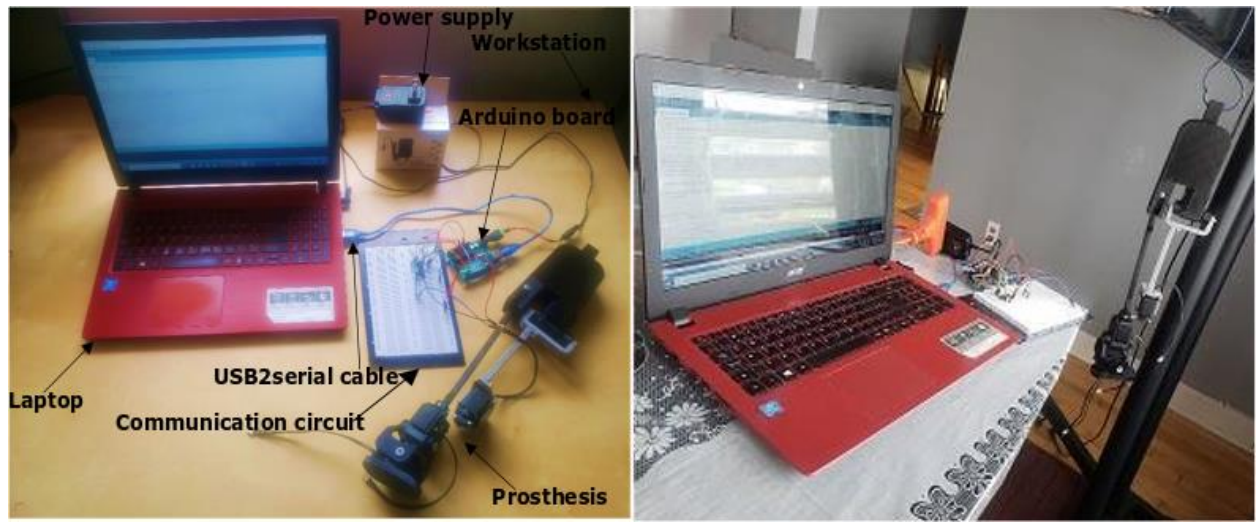

Figure 3 Experimental setup

The prosthetic joints were actuated thanks to Dynamixel AX-18A smart actuators controlled by the Arduino platform, chosen for their user-friendliness and simplicity. These servomotors were connected to an Arduino uno board (which integrated an ATMEGA328p microcontroller). To control the Dynamixel actuators, the main controller must convert its UART signals to the half-duplex type [55]. The communication circuit employed in this study is depicted in Appendix 1 and was based on the 74LS241 chip. The Arduino uno board was in turn connected via a USB cable to the computer for serial data transfer. The Dynamixel AX-18A actuators communicate with the Arduino according to the Serial protocol. When connected to the communication circuit and the Arduino uno board, these servomotors returned their feedback through the serial monitor of the Arduino.

### 2.3.2 Experimental protocol

The Dynamixel actuators were powered at 12V using a 9V-12V power source once the experimental setup was constructed and various equipment (Arduino board, PC, prototype, communication circuit) was properly linked. The physical prototype's joints were then asked to move independently using an algorithm control that controlled the servomotors at a specific speed and direction (i.e., in position control mode) [56].

Upper limb motions (elbow flexion–extension, forearm pronation–supination (PS), flexion–extension (FE), and radio-ulnar deviation of the wrist) were included in the protocol in order to reflect nontargeted upper limb movements that might occur in everyday life. However, because the efficiency/particularity of the prosthesis relies on the biofidelic mechanism of the forearm, the experimental tests focused on the forearm PS. Furthermore, the elbow movement is easy to implement.

Experimental tests were conducted in three sessions, during which the same tests were repeated at different times: the first and second sessions were conducted on the same day at a 5-minute interval, and 10 days elapsed between them and the third session. These tests aimed to measure torques and the range of motion (ROM) by obtaining from the Dynamixel actuators the feedback in position and in load for each joint through the IDE Arduino. During each session, the AX-18A servomotors for pronation–supination were programmed to move at a speed of 11.1 RPM in a counter-clockwise (CCW) direction for pronation and a clockwise (CW) direction for supination of the forearm. All positions allowed by the design were read using the Serial monitor of Arduino software for ROM measurement, with only the maximum positions from the neutral position (palm up) reported. The feedback in load was collected over a period of time under the same conditions (i.e., the same speed, directions, and motions) to allow the measurement of joint torques. For this second parameter, the start and finish positions, as well as the direction and speed, were sent to the servomotors in commands.

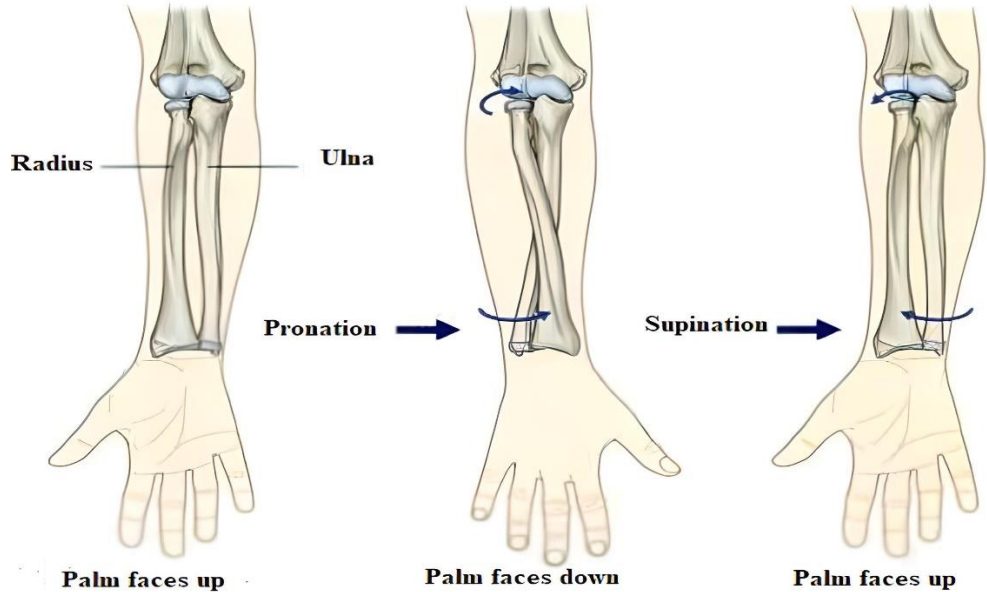

Figure 4 The forearm pronation–supination (PS). Adapted from [57].

### 2.3.3 Estimation of performance metrics: measurement of ROM and actuator torques

The range of motion (ROM) and torques were estimated using the feedback in position and in load, respectively, obtained from the Dynamixel servomotors, via the IDE Arduino.

The Dynamixel AX-18A actuators can operate between 0 and 300° in control position mode. The position feedback, on the other hand, is returned in 1024 bits with a unit of 0.29 degrees. As a result, the data read for the positions of Dynamixel actuators were multiplied by 0.29 degrees to convert them to joint positions.

Otherwise, the load feedback is returned in 2048 bits. The unit was 0.1 percent (0.1%) and the values read varied from 0 to 2047 (0-1023 for CCW direction and 1024-2047 for CW direction).

Through the relationship (9), the actuator's load (i.e., current) is proportional to the torque.

$$T = 1.8K \quad (9)$$

where K, the proportionality factor, is calculated by equations (10) and (11) from the load feedback:

- If **X**, the value of the load feedback, is less than 1023, i.e., **X** < 1023:

$$K = (X)\,0.1\% \quad (10)$$

- If **X**, the value of the load feedback, is more than 1023, i.e., X >1023:

$$K = (X - 1024)\, 0.1\% \qquad (11)$$

Concisely, joint torques are a percentage of the maximum torque (1.8 N.m). For example, a load feedback value of 256 indicates that the joint is controlled by 25% of the maximum torque, resulting in a torque of around 0.45 N.m.

### 2.3.4 Statistical Analysis

The system's effectiveness was measured using performance parameters such as range of motion (ROM) and joint torques, which were calculated and given as mean and standard deviation (SD), i.e., mean ± SD. In addition, the validity and reliability of experimental tests were assessed using the test–retest reliability method. Intra-session reliability (between two tests conducted on the same day at 5-minute intervals) and inter-session reliability (between the two first tests and a third test conducted 10 days later) were calculated for this purpose. The reliability coefficient (α) was used to interpret test reliability as follows: fair (below 0.60); moderate (0.60–0.69); acceptable (0.70–0.79); good (0.8–0.89); excellent (0.9–1.00).

## 3. RESULTS

This section presents the findings of multibody (MBD) simulations and experimental tests on the range of motion and joint torques during the pronation–supination (PS) task.

### 3.1 Simulation results of joint kinematics and dynamics

During the PS task, the range of motion was 146.84 ± 14.32°, ranging from -65.24 ± 7.16° to 81.60 ± 7.16°, according to the results of multibody dynamics (MBD) reported in Table 2. Minimum, maximum, and range of PS joint torques are also reported in Table 2. With minimum and maximum values of - 0.11 ± 0.02 N.m and 0.09 ± 0.03 N.m, respectively, the average PS torque range was approximatively 0.2 ± 0.05 N.m. Minimum and maximum values are different from one subject to another. However, the standard deviations around the average are relatively low for PS angles, and very low for joint torques. The highest maximum torque was 0.29N.m, as observed in subject No.12.

Table 2 Minimum (Min), maximum (Max) and range of pronation-supination (PS) angles, and torques during the PS task. The values are given as mean and standard deviation (SD) for the 15 subjects.

| Subjects No | Angles [°] | | | Joint torques [N.m] | | |
|---|---|---|---|---|---|---|
| | Min | Max | Range | Min | Max | Range |
| 1 | -54.27 | 78.79 | 133.06 | -0.1437 | 0.069 | 0.2127 |
| 2 | -78.58 | 83.62 | 162.2 | -0.076 | 0.08 | 0.1559 |
| 3 | -70.55 | 98.34 | 168.89 | -0.074 | 0.0348 | 0.1088 |
| 4 | -85.68 | 82.43 | 168.11 | -0.099 | 0.083 | 0.1820 |
| 5 | -71.05 | 85.4 | 156.45 | -0.09 | 0.0405 | 0.1305 |
| 6 | -80.66 | 79.31 | 159.97 | -0.088 | 0.069 | 0.1570 |
| 7 | -49.45 | 85.29 | 134.74 | -0.12 | 0.0538 | 0.1738 |
| 8 | -57.6 | 83.26 | 140.86 | -0.1539 | 0.089 | 0.2429 |
| 9 | -68.77 | 77.55 | 146.32 | -0.15 | 0.119 | 0.269 |
| 10 | -63.86 | 69.77 | 133.63 | -0.096 | 0.113 | 0.209 |
| 11 | -68.05 | 83.61 | 151.66 | -0.131 | 0.11 | 0.241 |
| 12 | -54.19 | 78.6 | 132.79 | -0.134 | 0.16 | 0.294 |
| 13 | -59.2 | 76.76 | 135.96 | -0.0815 | 0.096 | 0.1775 |

| 14 | -60.79 | 63.67 | 124.46 | -0.0895 | 0.11 | 0.1995 |
| 15 | -55.9 | 97.71 | 153.61 | -0.111 | 0.078 | 0.189 |
| Mean ± SD | -65,24 ± 7.16 | 81.60 ± 7.16 | 146.84 ± 14.32 | -0.11 ± 0.02 | 0.087 ± 0.03 | 0.197 ± 0.05 |

## 3.2 Experimental results

The kinematic results obtained from the experimental tests show that the physical prototype provides a forearm average rotation of up to 156.26 ± 4.71°, ranging from - 71.05 ± 2.35° to + 85.26 ± 2.35° during the pronation–supination (PS) task.

Experimental results of dynamics (represented as the PS joint torques of the main DoF) are reported in Table 3. The first session of measurements of joint torques (Trial 1) yielded a PS torque range of approximately 0.23 ± 0.05 N.m, with minimum and maximum values of - 0.03 ± 0.02 N.m and 0.2 ± 0.03 N.m, respectively. These measured torques present a relatively low standard deviation (from 0.02N.m to 0.03 N.m) during the PS task.

A second session (Trial 2) performed on the same day as the first at a 5-minute interval, and a third session (Trial 3) carried out 10 days later, led to PS torque ranges of 0.3 ± 0.05 N.m and 0.34 ± 0.08 N.m, respectively. As in Trial 1, the PS joint torques obtained from the second and third sessions still presented a low dispersion around the mean. The largest standard deviation around the average torque range was 0.08N.m. When compared to the multibody results, the experimental results show that the physical prototype increases the PS joint torques by 13%, 23%, and 41%, respectively, in the first, second, and third sessions.

Overall, there was a 40% difference between the average range of simulated torques and the average range of measured torques during the PS task, i.e., 0.2 N.m versus 0.28 N.m. However, with only a 3.4% difference, the maximum range of approximated torques, 0.29 N.m, was quite near to the average range of measured torques (0.28N.m).

**Reliability of experimental measurements**

To consider the reliability and validity of the experimental measurements, we compared the experimental measurements of three trials obtained by the same test over time. Intra-session reliability (between Trial 1 and Trial 2 conducted on the same day) and inter-session reliability (between the two first tests and a third test carried out 10 days later) were determined. Table 3 shows the detailed results for the reliability of experimental joint torques.

Intra-session reliability was excellent ($\alpha = 0.96\text{-}0.98$) for both pronation and supination. Inter-session reliability was excellent ($\alpha = 0.93$) for pronation, while good inter-session reliability was found during the supination task ($\alpha = 0.81\text{-}0.84$).

Table 3 Reliability of experimental measurements. The values of joint torque are given as mean and standard deviation (SD). Intra-session reliability is calculated between Trial 1 and Trial 2, and inter-session reliability between them and Trial 3.

|  | **Mean ± SD** | | **Mean ± SD** | **Reliability coefficient (α)** | | |
| --- | --- | --- | --- | --- | --- | --- |
|  | Trial 1 | Trial 2 | Trial 3 | Intra-session | Inter-session | |
|  | | | | Trial 1 & Trial 2 | Trial 1 & Trial 3 | Trial 2 & Trial 3 |
| Minimum/supination | -0.03 ± 0.02 N.m | -0.033 ± 0.015 N.m | -0.04 ± 0.05 N.m | 0.96 | 0.84 | 0.81 |
| Maximum/pronation | 0.2 ± 0.03 N.m | 0.225 ± 0.04 N.m | 0.3 ± 0.03 N.m | 0.98 | 0.93 | 0.93 |
| Range | 0.23 ± 0.05 N.m | 0.26 ± 0.05 N.m | 0.34 ± 0.08 N.m | | | |

| Average range | | 0.28 ± 0.06 N.m | |
|---|---|---|---|

## 4. DISCUSSION

The goal of this paper was to propose a test bench for evaluating upper limb exoskeletons quantitatively. This was accomplished by determining the range of motion (ROM) and joint torques using both kinematic modeling and experimental measurements (using sensors integrated into Dynamixel actuators, where ROM and joint torques were estimated from actuator feedback, respectively, in position and in load through the IDE Arduino).

### 4.1 Analysis of kinematics results

The results of multibody kinematics showed a range of motion (ROM) of approximately 146.84 ± 14.32°, while the experimental measurements provided a ROM average of 156.26 ± 4.71°. This means a difference of approximately 6.5% between the simulated ROM and the experimental values. We can conclude that the difference between the simulated ROM and the experimentally measured one is not significant, since, in experimental cases with more difficult realization conditions, a percentage deviation of up to 10% can be regarded as insignificant. When we considered the uncertainties in the comparison, we discovered that the uncertainty ranges for the two measurements overlapped between 151.55 ° and 160.97 ° (figure 5), implying that the real value of ROM may be somewhere between the simulated and experimental values.

This result demonstrates that the approximated and experimental values are close. When comparing measurements, we reached the conclusion that considering the uncertainty ranges is relevant.

We noted that this range of PS angles from 151.55 ° to 160.97 ° is not far from 150°, which can be considered a reference value for the pronation–supination of the forearm during daily activities. It is also consistent with the fact that the PS's typical ROM (i.e., the ROM of the healthy arm) is roughly 155° (70/85). As a result, the simulated ROM and those of the physical prototype, in addition to being close, fully cover the ROM required for upper limb activities of daily living (ADL), which is around 150° [58, 59].

These findings demonstrated that the procedures employed to generate these ROM values were relevant and helpful in achieving the study's goal.

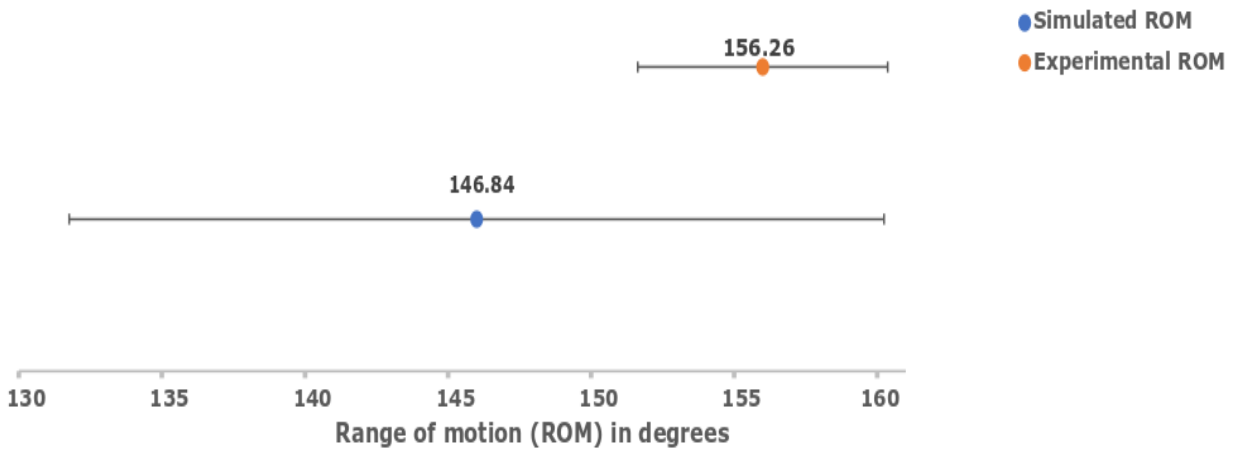

Figure 5 Uncertainty ranges of the simulated range of motion from multibody analysis (Simulated ROM) versus the one from experimental measurement (Experimental ROM).

## 4.2 Analysis of dynamics results

The dynamic results demonstrate the importance of considering the variations in torque (rather than average torque) in human joints during motion when sizing actuators for exoskeletons.

When compared to those calculated by multibody (MBD) analysis, the PS torque ranges measured experimentally were overestimated by up to 40% during the PS task (i.e., 0.20N.m vs 0.28 N.m). The highest torque range determined by multibody analysis, 0.29N.m, is, nevertheless, very close to the average range of measured torques (0.28N.m), with just a 3.4% difference. This can be explained by the fact that exoskeletons are typically designed on the basis of the maximal joint torques and velocities, due to the variations in the torques and velocities in human joints during motion. When we consider the evolution of power over time, as shown in Figure 6, we can identify the power peaks, confirming that there are torque fluctuations in PS joints throughout motion.

These findings prove that the maximum torque range (rather than the average) obtained from MBD simulations is important for actuator sizing for the PS joint. When we compare this maximum torque range and experimental torque ranges (i.e., the average from three trials), it leads to a low difference (3.4%), which indicates that the highest range of estimated joint torque estimated through MBD is close to the torque range measured experimentally (i.e., thanks to the physical prototype). As a result, we can conclude that multibody analysis has the potential to be a useful technique in the quantitative evaluation of upper limb exoskeletons. This study also highlights the importance of sizing actuators based on maximum joint torques.

Regarding the experimental method (based on the use of sensors integrated into Dynamixel actuators), all three experimental trials presented a relatively low standard deviation (SD = 0.05-0.08), which indicates a low dispersion of measured torques around the average and thus a low variability between the measurements of PS joint torques.

The purpose of repeating the experimental test for the measurement of joint torque was to examine the reliability of the experimental technique for the quantitative evaluation of upper limb exoskeletons suggested in this work.

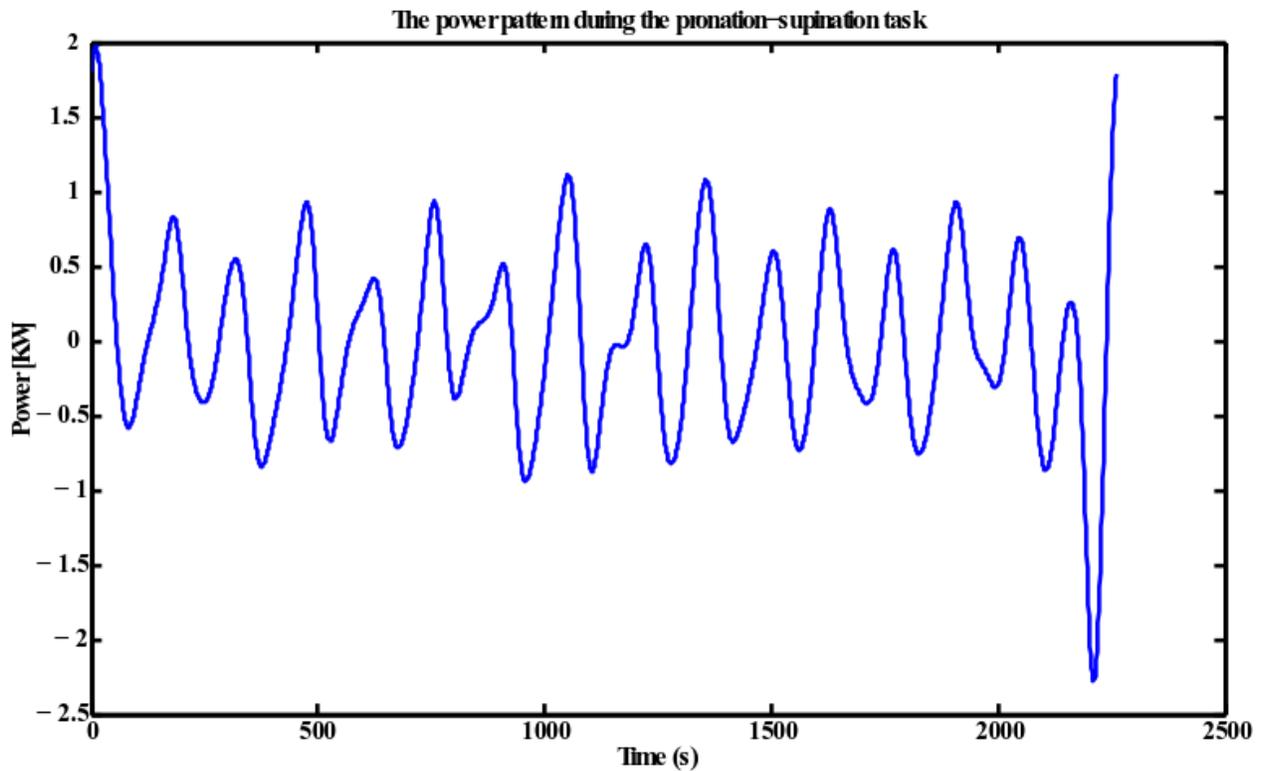

Figure 6 The power pattern during the pronation-supination task

## 4.3 Reliability of experimental measurements

Within the sessions, the experimental results were extremely reliable (α = 0.96-0.98), and they were highly or very reliable between the sessions (α > 0.85). This demonstrates that the measurement tool chosen for the experimental tests could be a viable assessment method for upper limb exoskeletons.

Intra-session reliability was excellent for both pronation and supination (α = 0.96-0.98). Excellent test–retest reliability indicates the internal validity of the test and ensures that the measurements obtained for ROM and joint torques are both representative and stable over time.

Inter-session reliability was excellent for pronation (α = 0.93), whereas good inter-session reliability was found during the supination task (α = 0.81- 0.86). Good reliability means that the parameters (ROM and joint torques) were appropriately measured and that the same results could be reliably reproduced in the same situation and under the same conditions.

## 4.4 Limitations and perspectives of the study

The following two major limitations should be considered when interpreting the findings of this study. To begin, experimental tests were conducted only with the device without subjects. Then, within this study, ROM and joint torques were considered in order to evaluate the performance or effectiveness of the system in order to facilitate the interpretation of the results. While these parameters are relevant and the evaluation appears promising, the safety aspect was not taken into account in the effectiveness/performance evaluation of the system, which represents a further limitation. Therefore, the safety aspect is seen as a market perspective rather than a limit [46]. Future research would require the performance of experimental tests not only with the system but also with human subjects in order to consider the user's safety or undesired effects of the exoskeleton by quantifying the kinematic coupling between user and device. It is known that a compliant kinematic structure contributes to reducing the risk of injury among exoskeleton users [34] and, as a result, deals with the safety concern in the exoskeleton design. The introduction of kinematic coupling as an additional parameter may improve the approach to the evaluation of upper limb exoskeletons suggested in this work. Furthermore, quantifying the kinematic coupling will allow for a more thorough evaluation, especially since the test bench employed in this work has a distinguishing design feature: its biofidelic kinematic structure.

Due to the simplification goal and because the elbow movement is simple to accomplish, only one type of mobility, i.e., the pronation–supination of the forearm, was taken into consideration. While the test bench's efficiency and its particularity are based on the forearm's biofidelic mechanism, it is necessary to assess the elbow's flexion–extension in order to validate the suggested approach for more dynamic tasks. A similar approach might also be applied to other anatomical joints or limbs, as well as to multi-joint exoskeletons.

## 5. CONCLUSION

A test bench utilizing a biofidelic prosthesis developed for evaluating upper limb exoskeletons is established and assessed. In the quantitative evaluation of the system, both kinematic modeling and experimental measurements (which rely on sensors integrated into Dynamixel actuators) are used. The system's performance is evaluated by range of motion (ROM) and joint torques during pronation–supination (PS) tasks. When assessing the uncertainty ranges of ROM and the maximal torque range for the MBD results, these parameters are considered relevant since similar results are obtained between numerical (i.e., multibody (MBD)) and experimental methods.

The current study demonstrates the importance of considering the variations in torque (rather than average torque) in human joints during motion when sizing actuators for exoskeletons. The findings prove that the maximum torque range (rather than the average) obtained from MBD simulations is important for actuator sizing for the PS joint. This study also highlights the importance of sizing actuators based on maximum joint torques. In addition to being close, the range of motion (ROM) generated by the MBD model and that measured from the prototype fully cover the ROM required for upper limb activities of daily living (ADL), which is roughly 150°.

Finally, multibody analysis has the potential to be a valuable technique for the quantitative evaluation of upper limb exoskeletons. The measurement tool adopted for experimental tests could be a reliable and replicable assessment approach for upper limb exoskeletons in terms of an experimental technique. The experimental results were extremely reliable within the sessions ($\alpha = 0.96\text{-}0.98$) and highly ($\alpha = 0.93$) or very reliable ($\alpha = 0.81\text{-}0.86$) between sessions.

The proposed test bench could become an important, easy-to-use tool for evaluating exoskeletons for the upper limb, and a similar approach could be extended to other anatomical joints or limbs. However, while promising, future research would require the performance of experimental tests not only with the system but also with human subjects in order to consider the user's safety or undesired effects of the exoskeleton by quantifying the kinematic coupling between the user and the device.

## 6. CONFLICT OF INTEREST

The authors declare that the research was conducted in the absence of any commercial or financial relationships that could be construed as a potential conflict of interest.

## 7. FUNDING

## 8. ACKNOWLEDGMENTS

We thank Fredérick Leclerc for her help in the fabrication of prosthesis prototype by 3D printing.

## 9. APPENDIX

**Appendix 1:**

Figure 7 The communication circuit of DYNAMIEL AX-18A servomotor based on the 74LS241 chip. To control Dynamixel, main Controller and DYNAMIXEL communicate each other thanks to this circuit by sending and receiving data, respectively by TXD and RXD pins. In this circuit, Arduino board should be supply at 5V (VCC) and each Dynamixel between 9V-12V (Vin).

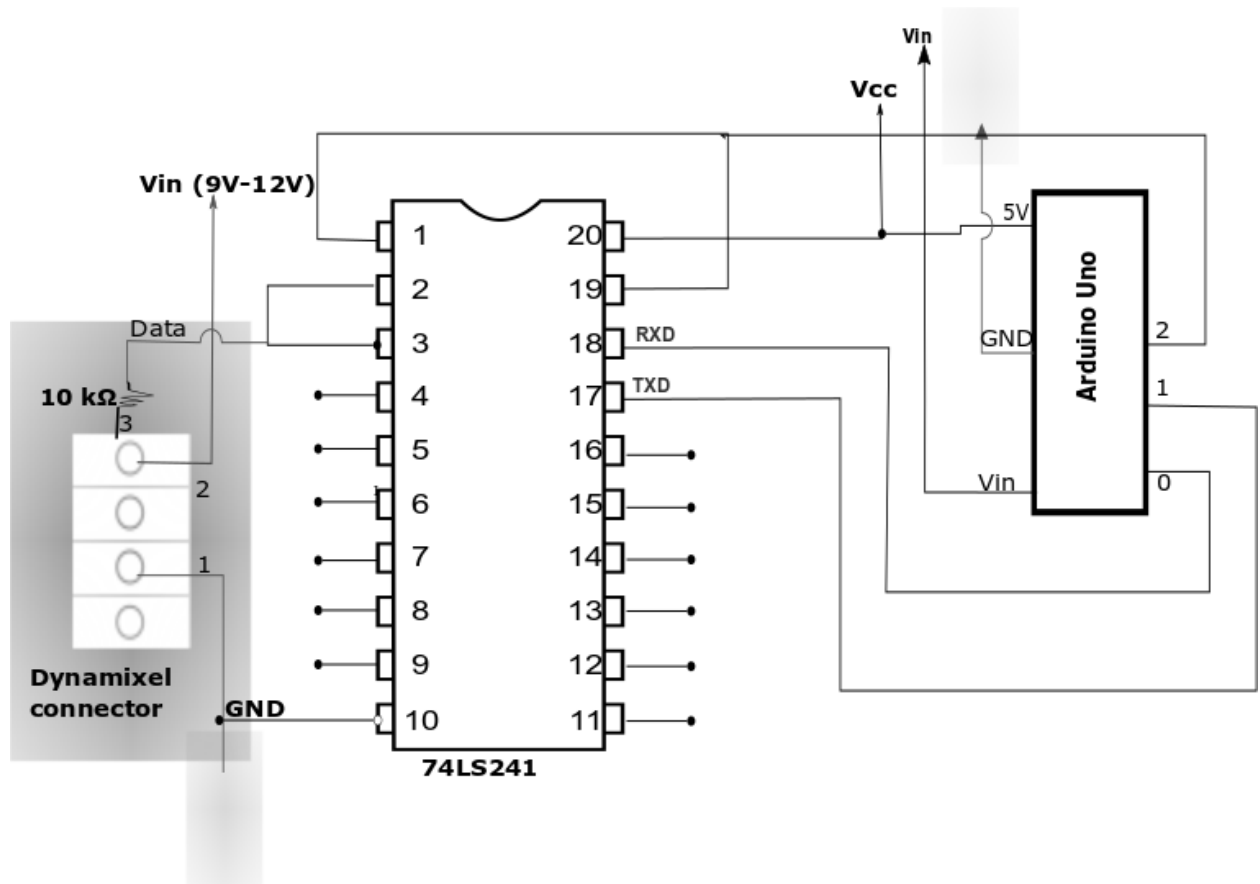